\documentclass[10pt,twocolumn,letterpaper]{article}

\usepackage{iccv}              

\newcommand{\sharedfootnotetext}{%
  \hypertarget{sharedfn}{}%
  \footnotetext[1]{P-values reflect the likelihood of observing results by random chance; smaller values suggest stronger associations.}%
}
\newcommand{\sharedfootnotemark}{\protect\hyperlink{sharedfn}{\textsuperscript{*}}}

\definecolor{iccvblue}{rgb}{0.21,0.49,0.74}
\usepackage[pagebackref,breaklinks,colorlinks,allcolors=iccvblue]{hyperref}
\hypersetup{
    colorlinks=true,
    allcolors=blue,
}

\usepackage{amsmath,amsthm,amssymb}
\usepackage{makecell} 
\usepackage{mathtools}
\usepackage{tikz}
\usepackage{wrapfig}
\usepackage{graphicx}
\usetikzlibrary{positioning}
\usepackage{caption}
\usepackage{subcaption}
\usepackage[accsupp]{axessibility}  

\newtheorem{definition}{Definition}
\newtheorem{theorem}{Theorem}

\def\paperID{14695} 
\def\confName{ICCV}
\def\confYear{2025}

\title{The Inter-Intra Modal Measure: A Predictive Lens on Fine-Tuning Outcomes in Vision-Language Models}

\author{Laura Niss\thanks{Equal contribution.} \quad Kevin Vogt-Lowell\textsuperscript{*} \quad Theodoros Tsiligkaridis\\
{\tt\small \{laura.niss, kevin.vogt-lowell, ttsili\}@ll.mit.edu}\\
MIT Lincoln Laboratory\\
244 Wood St, Lexington, MA 02421
}

\begin{document}
\maketitle
\begin{abstract}
The fine-tuning of large vision-language foundation models remains an underexplored area, particularly regarding its impact on learning gains and catastrophic forgetting. Inspired by the significance of modality gaps in contrastive dual-encoders, we introduce the Inter-Intra Modal Measure (IIMM)—a predictive metric that quantifies the relationship between intra-modal image embedding similarity and inter-modal misalignment. Through extensive empirical analysis across four state-of-the-art vision-language models and five fine-tuning techniques, we establish a strong linear relationship: tasks with higher IIMM scores yield greater in-domain performance improvements but suffer from more pronounced out-of-domain degradation, with some parameter-efficient fine-tuning (PEFT) methods exhibiting severe forgetting. Compared to existing transferability measures, the IIMM demonstrates significantly stronger predictive power for accuracy changes post fine-tuning in dual-encoder models. Moreover, we provide a theoretical bound, proving that changes in IIMM are limited by the Wasserstein distance between pre- and post-fine-tuning embedding distributions, ensuring its stability and robustness as a predictive measure. With only a single forward pass of the target data, practitioners can leverage this key insight to evaluate the degree to which a model can be expected to improve following fine-tuning. When combined with prior knowledge of a model’s performance across diverse tasks, the IIMM further enhances transferability predictions for novel tasks, offering a lightweight yet effective tool for guiding model adaptation strategies. Our code is provided at \href{https://github.com/mit-ll/IIMM}{https://github.com/mit-ll/IIMM}.
\end{abstract}    
\section{Introduction}
\label{sec:intro}

\begin{figure*}[ht]
    \begin{center}
        \includegraphics[width=0.74\textwidth]{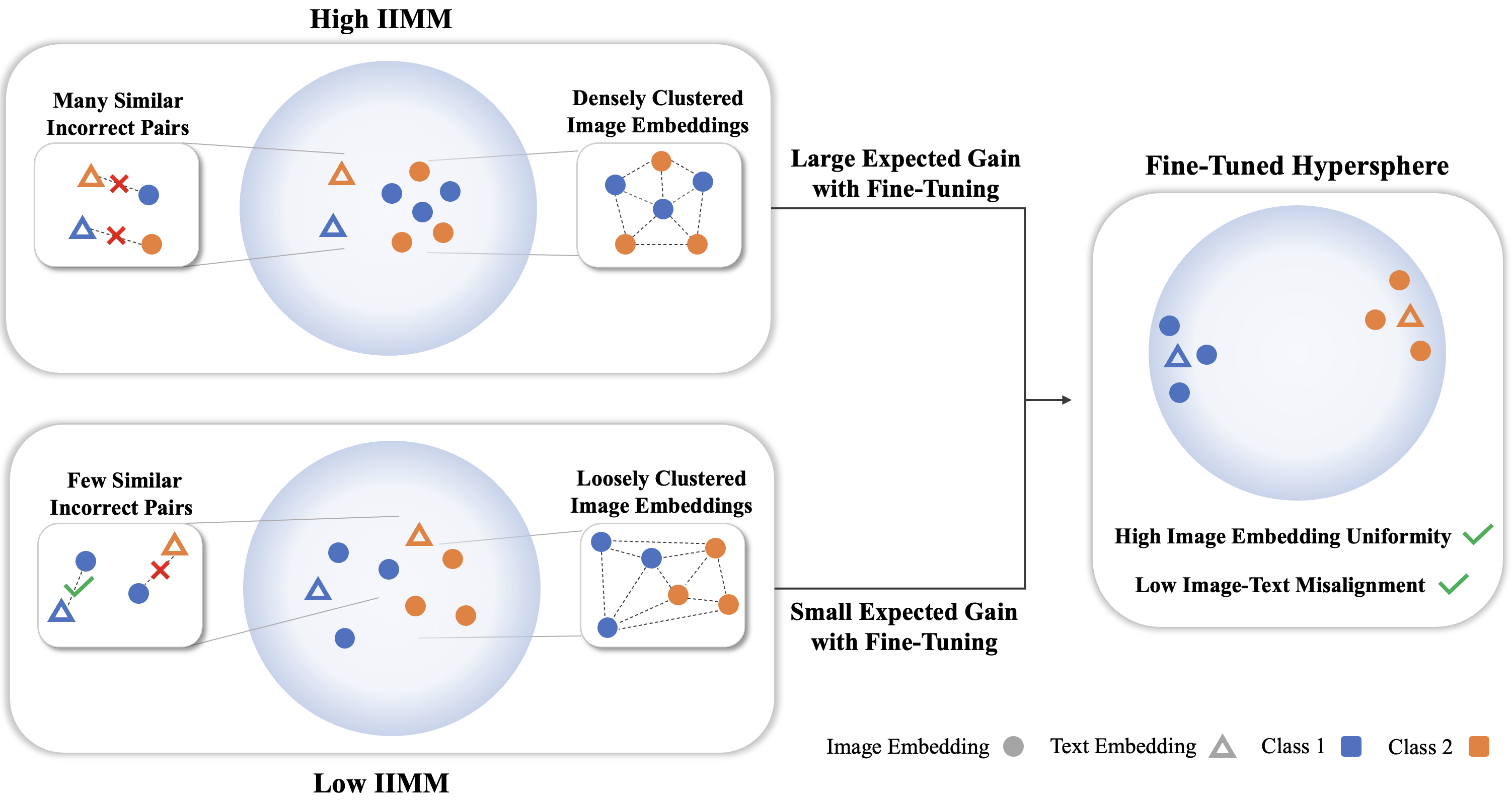}
    \end{center}
    \caption{The Inter-Intra Modal Measure (IIMM) combines two key aspects of embedding geometry: intra-modal image embedding similarity, and inter-modal alignment between image and text embeddings. This measure captures information about the embedding geometry that demonstrates a strong relationship to performance gains observed with fine-tuning.}
    \label{fig:problem_setup}
\end{figure*}

Large vision-language foundation models have demonstrated superior zero-shot generalization compared to traditional vision models, leveraging web-scale pre-training to achieve state-of-the-art performance across diverse tasks without task-specific supervision \citep{radfordLearningTransferableVisual2021a}. Advances in enhancing zero-shot generalization have also been proposed \citep{Menon:2023, Roth:2023, Yu:2025}. Fine-tuning can further enhance performance on specific downstream applications \citep{bommasani2021opportunities, zhang2025parameter, Liao:DWSoups:2024, Teterwak:2025, Liao:MUDG:2025}. However, fine-tuned foundation models still struggle on certain tasks, particularly in specialized fields such as satellite and medical image classification \citep{awaisFoundationalModelsDefining2023,huixAreNaturalDomain2024, xieWhenAreFoundationModelsEffective2024, vogt2023robust}. Given the computational cost of fine-tuning, determining whether adaptation will lead to meaningful improvements remains a crucial challenge.

Beyond resource efficiency, fine-tuning also introduces the risk of catastrophic forgetting—where a model loses previously learned knowledge while improving on the new task. This effect is particularly pronounced in dual-encoder vision-language models, where fine-tuning disrupts the alignment between image and text embedding spaces, degrading the model’s ability to generalize to out-of-domain tasks \citep{dingDonStopLearning2022, niContinualVisionLanguageRepresentation2023}. This creates an inherent trade-off between learning and forgetting, making it critical for practitioners to anticipate the effects of fine-tuning before allocating computational resources \citep{yuBoostingContinualLearning2024, zhengPreventingZeroShotTransfer2023a, mukhoti2023fine}. A predictive metric that captures both learning gains and forgetting risks would provide a valuable tool for optimizing fine-tuning decisions.

The challenge of estimating a model’s fine-tuned performance a priori falls under the domain of model transferability \citep{WangNCTI2023, pandyTransferabilityEstimationUsing2022}. Common transferability measures assess a pre-trained model’s suitability for a given task by analyzing its latent representations. However, most traditional methods rely on source data and task labels, making them incompatible with contrastively trained foundation models that lack explicit supervision \citep{shaoEMMS2024}. Furthermore, newer transferability metrics are designed for uni-modal settings, failing to account for inter-modal interactions in vision-language models. Among approaches that estimate post-fine-tuning performance, evaluation is often limited to rank correlation rather than directly predicting performance changes \cite{shaoSFDA2022, huangFrustratinglyEasy2022}.

\paragraph{Present work}

To address these limitations, we introduce the Inter-Intra Modal Measure (IIMM), a novel metric that quantifies both intra-modal representation spread and inter-modal alignment in contrastive vision-language models, capturing the fundamental trade-off between learning and forgetting without the need for fine-tuning. We compare our measure against several standard and recent transferability measures in both rank correlation and strength of predictive fit to performance gains after fine-tuning. To evaluate our measure, we perform fine-tuning across nine well-known vision classification tasks using four state-of-the-art vision-language models and five fine-tuning techniques, analyzing in-domain improvements as well as induced catastrophic forgetting across eight out-of-domain tasks. We compare our measure against six transferability measures applicable to vision-language encoders, and show that our measure has a significantly stronger monotonic relationship with both fine-tuned in-domain accuracy and in-domain performance gains scaled by the zero-shot error. When fitting a simple linear regression of the IIMM to the scaled in-domain accuracy gains, we achieve $R^2>0.85$ for all base models. This offers a practical and theoretically grounded tool for model adaptation in vision-language tasks.

\paragraph{Contributions}
In summary, our work provides the following contributions:
\begin{enumerate}
    \item We introduce the Inter-Intra Modal Measure (IIMM), a standalone metric that integrates inter–modal misalignment and intra–modal uniformity to predict fine–tuning outcomes in dual–encoder vision–language models.
    \item We derive the IIMM directly from contrastive learning principles, establishing a clear connection between the geometry of the embedding space and transferability estimation.
    \item We prove a theoretical bound showing that changes in the IIMM are constrained by the 1–Wasserstein distance between pre– and post–fine–tuning embedding distributions. This result guarantees that  significant shifts in the representation space will lead to substantial changes in IIMM, underscoring its stability as a predictive measure.
    \item We show that the zero-shot IIMM has a strong, positive linear relationship with learning and a negative relationship with forgetting post fine-tuning. We also provide empirical evidence that for any vision-language dual-encoder, fine-tuning on only a few standard vision benchmarks is sufficient to fit a strong linear predictor of fine-tuning accuracy gains with the IIMM for transfer estimation of novel tasks.
\end{enumerate}
\section{Related work}

\paragraph{Dual-encoders and the modality gap} Multi-modal dual-encoders trained using contrastive learning consist of two models jointly trained to map each modality into a shared representation space \citep{radfordLearningTransferableVisual2021b,jiaScalingVisualVisionLanguage2021,liAlignFuseVision2021,xuVideoCLIPContrastivePretraining2021,zhangContrastiveLearningMedical2022}. Previous research has highlighted the existence of a modality gap within this joint embedding space that results from the encoders initializing embeddings within random cones, and that contrastive loss maintains this initialized gap \citep{liangMindGapUnderstanding2022}. Additionally, empirical evidence shows that  manipulating this gap can impact zero-shot performance. While catastrophic forgetting has been proven to significantly affect traditional supervised learning techniques \citep{mccloskeyCatastrophicInterferenceConnectionist1989, kirkpatrickOvercomingCatastrophicForgetting2017}, recent research has shown that self-supervised, uni-modal models do not suffer from such severe forgetting \citep{ni2021selfsupervised, hu2022does}. Yet, with multi-modal foundation models, such as CLIP, continuous training can cause misalignment of the image and text embeddings, an outcome highlighted and explored in several works \citep{dingDonStopLearning2022,linSpecialityVsGenerality2023,niContinualVisionLanguageRepresentation2023,NEURIPS2022_bd361197,fan2022unified}.

\paragraph{Transferability estimation}
A significant amount of research has focused on addressing transferability estimation with respect to uni-modal encoders. Estimation of model transferability often involves transferability metrics, or measures that predict how well knowledge learned from source tasks will transfer to new target tasks. \citep{ding2024model} The approaches for quantifying these metrics vary greatly, with some approaches simply defining them as the log-likelihood of a target task. Methods that rely upon a log-likelihood definition are inherently well-suited for model selection because the definition relies directly upon the size of the target dataset, and within model selection the target dataset is held constant. Regardless of approach, nearly all metrics are evaluated via rank correlation across a model set \cite{shaoEMMS2024, shaoSFDA2022, WangNCTI2023, huangFrustratinglyEasy2022}.

Some modern approaches include training small neural networks and comparing the similarity of resultant task embeddings \citep{dwivediRepresentationSimilarityAnalysis2019}, training a full neural network on large amounts of models trained across many tasks to learn a similarity score function \citep{liGuidedRecommendationModel2023}, and comparing source training data to new target task data to estimate similarity \citep{cuiLargeScaleFineGrained2018}. Some classic methods, such as LEEP \citep{nguyenLEEPNewMeasure2020} and NCE \citep{tranTransferabilityHardnessSupervised2019}, rely upon information from the source training data and are thus inapplicable to self-supervised foundation models; they also cannot be applied to dual-encoders. However, there are a range of uni-modal methods that can be easily applied to the vision encoder of a dual-encoder model \citep{youLogme2021,huangFrustratinglyEasy2022,shaoEMMS2024}. 

For multi-modal models, some recent research has involved zero-shot model selection using language to infer the transferability of a dataset, such as LOVM \citep{zoharLOVMLanguageOnlyVision2023} and SWAB \citep{ yiBridgeModalityCapacity2024}. These methods do not claim to estimate model performance after fine-tuning, and are specific to the zero-shot setting.

\section{Inter-Intra Modal Measure} \label{sec:IIMM}
In this section, we provide a principled derivation of the IIMM metric from contrastive learning principles. In addition, we prove a theoretical bound and provide intuition behind the metric for predicting fine-tuning outcomes.

\paragraph{Notation}
Let \(t\) denote a target task, and let \(\mathcal{T}\) be the set of tasks under evaluation. Consider a training dataset for task \(t\) composed of image–text pairs. Let $X$ denote the set of image embeddings, where each \(x \in X \subset \mathbb{R}^d\). Similarly, let $Y$ denote the set of text embeddings, where each \(y \in Y \subset \mathbb{R}^d\). For every image \(x\), let \(y(x) \in Y\) be the text embedding corresponding to its correct label.

Let \(\mu\) and \(\nu\) be two probability measures on a metric space \((\mathcal{X}, d)\). The 1–Wasserstein distance between \(\mu\) and \(\nu\) is defined as:
\[
W_1(\mu, \nu) = \inf_{\gamma \in \Gamma(\mu,\nu)} \int_{\mathcal{X} \times \mathcal{X}} d(x,y) \, d\gamma(x,y),
\]
where \(\Gamma(\mu,\nu)\) denotes the set of all couplings (i.e., joint distributions) with marginals \(\mu\) and \(\nu\).

\subsection{Derivation via Contrastive Learning Principles}
We begin with the widely used InfoNCE loss in contrastive learning \citep{oord2018representation}. For an image embedding \(x \in X\) and its corresponding (correct) text embedding \(y \in Y\), the loss is defined as
\begin{equation}
L(x,y,\{y'\}) = -\log \frac{\exp\left(\frac{x^\top y}{\tau}\right)}{\exp\left(\frac{x^\top y}{\tau}\right) + \sum_{y' \in Y \setminus \{y(x)\}} \exp\left(\frac{x^\top y'}{\tau}\right)},
\label{eq:infonce}
\end{equation}
where \(\tau > 0\) is the temperature parameter and the summation is over all incorrect text embeddings \(y'\) (i.e. \(y' \neq y(x)\)). Minimizing the InfoNCE loss encourages two main effects:
\begin{enumerate}
    \item \textbf{Positive Pair Alignment:} The numerator is maximized when the cosine similarity \(x^\top y\) is high.
    \item \textbf{Negative Pair Separation:} The denominator penalizes high similarities between \(x\) and any incorrect text embedding \(y'\), thus pushing down \(x^\top y'\) for \(y' \neq y(x)\).
\end{enumerate}
The resulting changes in embedding geometry can be decomposed into two components: inter-modal misalignment and intra-modal uniformity.
\paragraph{Inter–modal misalignment}  
The effect on incorrect pairs can be quantified by the average similarity between an image embedding \(x\) and the set of incorrect text embeddings:
\begin{equation}
S_{\text{inter}} = \frac{1}{|X|} \sum_{x \in \mathcal{X}} \frac{1}{|Y|-1} \sum_{y' \in Y\setminus\{y(x)\}} x^\top y'.
\label{eq:inter_modal}
\end{equation}
A high \(S_{\text{inter}}\) indicates that the model has not sufficiently separated the image embedding from non-matching text embeddings. While indicative of potential for improvement, this can also be problematic, as fine-tuning may lead to over-alignment and subsequent catastrophic forgetting.

\paragraph{Intra–modal uniformity}  
In addition to inter–modal alignment, contrastive learning also encourages a good spread, or uniformity, of embeddings within each modality \citep{wangUnderstandingContrastiveRepresentation2020}. For image embeddings, we quantify uniformity by computing the average pairwise similarity:
\begin{equation}
S_{\text{intra}} = \frac{2}{|X|(|X| - 1)} \sum_{1 \le i < j \le |X|} x_i^\top x_j.
\label{eq:intra_modal}
\end{equation}
A high \(S_{\text{intra}}\) implies that the image embeddings are overly clustered, suggesting that there is significant potential for further separation (or re–organization) during fine–tuning.

\paragraph{Formulation and intuition}
Based on the above principles, we define the Inter–Intra Modal Measure (IIMM) as the average of the inter–modal and intra–modal terms:
\begin{equation}
\text{IIMM} = \frac{1}{2}\left( S_{\text{inter}} + S_{\text{intra}} \right).
\label{eq:iimm}
\end{equation}
A \textit{high} IIMM value indicates that the average similarity between an image and incorrect text embeddings is high and/or the image embeddings are highly clustered. This situation suggests that while there is significant capacity for learning (i.e., re–aligning the modalities), the risk of catastrophic forgetting is also high. Conversely, a \textit{low} IIMM value implies that the image and text modalities are well separated (low inter–modal similarity for negatives) and the image embeddings are sufficiently dispersed. In this case, fine–tuning is expected to yield minimal performance changes.

The IIMM metric arises naturally from the objectives of contrastive learning:
\begin{enumerate}
    \item The InfoNCE loss drives positive pairs to be closely aligned and negative pairs to be well separated.
    \item The average similarity over incorrect pairs, \(S_{\text{inter}}\), captures the degree of inter–modal misalignment.
    \item The average similarity among image embeddings, \(S_{\text{intra}}\), measures intra–modal uniformity.
\end{enumerate}
By combining these two measures as in Equation~\eqref{eq:iimm}, the IIMM provides a principled, predictive metric that reflects both the learning potential and the risk of catastrophic forgetting in vision–language models.

We provide further exploration of this measure and its relation to learning, forgetting, and the uniformity of the embedding space in \cref{sec:measure derivation}. Additionally, \cref{tab:cc base} provides an analysis of various convex combinations of the intra- and inter-modal sub-scores, highlighting modification of component weights as a means of evaluating each component's contributions to the predicted outcomes.

\subsection{Theoretical Bound}
In this section, we provide a theoretical result that links the change in IIMM to the Wasserstein distance between the joint distributions of incorrect image--text pairs (and image--image pairs) before and after fine-tuning. This bound offers a rigorous guarantee: if fine-tuning only induces a small distributional shift as measured by the Wasserstein distance, then the IIMM metric will change by no more than a proportional amount. Such a result is important as it justifies the use of IIMM as a robust indicator—ensuring that minor, perhaps noisy, changes in the embeddings do not lead to erratic changes in the IIMM, and that significant changes in the IIMM are indeed indicative of a meaningful shift in the model's representation.

\begin{theorem} \label{thm:IIMM_bound}
Assume that all image and text embeddings lie on the unit sphere $S^{d-1} \subset \mathbb{R}^d$. Let the \emph{IIMM} metric be defined by
\[
\operatorname{IIMM}(P) = \frac{1}{2}\Bigl( A(P) + B(P) \Bigr),
\]
where
\[
A(P) = \mathbb{E}_{(x,y') \sim P_{XY}} \bigl[x^\top y'\bigr],
\]
with $P_{XY}$ being the joint distribution of \emph{incorrect} image--text pairs, and
\[
B(P) = \mathbb{E}_{(x,x') \sim P_{X}\otimes P_{X}} \bigl[x^\top x'\bigr],
\]
with $P_{X}$ being the marginal distribution of image embeddings. Consider perturbed versions of these distributions $Q_{XY}$ and $Q_{X}$, respectively. Assume that the $1$-Wasserstein distances satisfy
\[
W_1(P_{XY}, Q_{XY}) \le \delta_A \quad \text{and} \quad W_1(P_{X}, Q_{X}) \le \delta_B.
\]
Then the following bound holds:
\[
\Bigl| \operatorname{IIMM}(Q) - \operatorname{IIMM}(P) \Bigr| \le \frac{\delta_A}{2} + \delta_B.
\]
Furthermore, if one computes empirical estimates $\widehat{\operatorname{IIMM}}(P)$ and $\widehat{\operatorname{IIMM}}(Q)$ using $N$ independent samples, then with probability at least $1-\eta$,
\[
\Bigl| \widehat{\operatorname{IIMM}}(Q) - \widehat{\operatorname{IIMM}}(P) \Bigr| \le \frac{\delta_A}{2} + \delta_B + \epsilon_N,
\]
where 
\[
\epsilon_N = O\!\Bigl(\sqrt{\frac{\log(1/\eta)}{N}}\Bigr)
\]
accounts for the sampling error.
\end{theorem}
Our empirical results have indicated that IIMM is strongly correlated with performance gains and catastrophic forgetting. The bound provided here offers a theoretical explanation: only significant reorganization of the embedding space (reflected by a large Wasserstein distance) can cause a substantial shift in IIMM. Therefore, a high IIMM change can be interpreted as evidence of a considerable shift in the model's internal representation, likely correlating with notable performance changes.

\section{Experiments}
\label{sec:experiental results}

\paragraph{Datasets}
To evaluate the relationships between transferability measures and model performance changes following fine-tuning, we compiled a diverse collection of nine commonly used, publicly available computer vision benchmarks provided by PyTorch \citep{liangMindGapUnderstanding2022,zoharLOVMLanguageOnlyVision2023,radfordLearningTransferableVisual2021a} as our testing dataset, with the goal of minimizing domain overlap and covering a range of coarse and granular classification tasks. These datasets included Stanford Cars \citep{krause3DObjectRepresentations2013}, CIFAR100 \citep{krizhevskyLearningMultipleLayers}, the Describable Textures Dataset (DTD) \citep{cimpoiDescribingTexturesWild2014}, satellite imagery datasets EuroSAT \citep{helberIntroducingEurosatNovel2018} and RESISC45 \citep{chengRemoteSensingImage2017}, the German Traffic Sign Recognition Benchmark (GTSRB) \citep{stallkampManVsComputer2012}, MNIST \citep{lecunGradientbasedLearningApplied1998}, the Scene Understanding Benchmark (SUN397) \citep{xiaoSUNDatabaseLargescale2010}, and the Street View House Numbers (SVHN) dataset \citep{netzerReadingDigitsNatural2011}. For each dataset, we generated validation splits by randomly sampling 10\% of the respective training data, capping the validation split size at 5000 samples for larger datasets.

\paragraph{Models and fine-tuning methods}

To investigate the applicability of IIMM to vision-language encoders, we conducted all experiments using two distinct groups of evaluations schema: one in which the base pre-trained model varies while the fine-tuning strategy is held constant, and another group in which the base pre-trained model is held constant while the fine-tuning strategy varies. We will refer to these evaluation schema as the base model evaluation and PEFT evaluation, respectively.

For the base model evaluation, we perform full fine-tuning over all parameters for the following four pre-trained models: CLIP, EVA-02-CLIP \citep{fang2023eva}, SigLIP \citep{Zhai_2023_siglip}, and CoCa \citep{yu2022coca}, using a ViT-B/32 image backbone for CLIP and CoCa and a ViT-B/16 image backbone for SigLIP and EVA-02-CLIP. 

For the PEFT evaluation, we utilized a CLIP model with a ViT-B/32 image backbone as the base model and evaluate four common parameter-efficient fine-tuning (PEFT) methods: CLIP-Adapter \citep{gaoCLIPAdapterBetterVisionLanguage2021}, BitFit (bias tuning) \citep{benzakenBitFitSimpleParameterefficient2022}, LoRA \citep{huLoRALowRankAdaptation2022a}, and a method we call Attention-Weight Tuning, which simply refines the fine-tuning of all attention layer weights in \citet{touvronThreeThingsEveryone2022} to just the attention layers' in-projection weights. 

\paragraph{Baselines}

While no methods currently exist for explicitly predicting gains in accuracy from fine-tuning for dual-encoder vision-language models, many related methods exist for stand-alone vision encoders. In particular, we compare against several recent or established measures self-described as transferability estimators. The six comparison measure are: TransRate \citep{huangFrustratinglyEasy2022}, 
Self-challenging Fisher Discriminant Analysis (SFDA) \citep{shaoSFDA2022}, Neural Collapse informed Transferability Index (NCTI) \citep{WangNCTI2023},  Efficient Multi-task Model Selector EMMS \citep{shaoEMMS2024}, Gaussian Bhattacharyya Coefficient  (GBC) \citep{pandyTransferabilityEstimationUsing2022}, and H-score \citep{baoInformationTheoreticApproachTransferability2019}.

We chose these comparison measures to span the definitions and applications of self-described transferability measures. Those that estimate the log-likelihood are H-Score, SFDA, TransRate, and EMMS. NCTI functions as an estimator of the gap between the embedding geometry of the pre-trained model and the embedding geometry of a hypothetical fine-tuned model \citep{WangNCTI2023}. The GBC quantifies transferability as a measure of overlap in the feature space between embedding clusters of different class labels \citep{pandyTransferabilityEstimationUsing2022}. 

\paragraph{Evaluation measures}
We evaluate performance of the transferability measures through the strength of their relationship to three variables: accuracy on the target test data after fine-tuning on the target task training data; gain in performance from the pre-trained model to the fine-tuned model scaled by the pre-trained model error, which we refer to as gain over zero-shot error; and average accuracy across the remaining eight non-target datasets to measure loss in performance from the pre-trained model on non-target tasks.

\begin{definition}[Gain over zero-shot error]
For a given task, we define $A_{zs}$ as the zero-shot accuracy and $A_{ft}$ as the fine-tuned accuracy. We then define the gain over zero-shot error as,

\begin{align}
    \frac{A_{ft} - A_{zs}}{1-A_{zs}}
\end{align}

This value normalizes for differences in the zero-shot accuracy across models and tasks.
\end{definition}

We first evaluate all measures' monotonic relationship to both fine-tuned accuracy and gain over zero-shot error using Kendall's $\tau$, which is a standard evaluation measure within the literature. We then look at the top performing measures of this rank correlation and fit a least-squares model to evaluate their predictive strength, additionally looking at a least-squares model fit of the top measure against the average accuracy on off-target tasks.

\begin{table*}[h!]
    \small
    \centering
    \begin{tabular}{lll|ll|ll|ll}
        \toprule
        \multicolumn{1}{c}{Model}& \multicolumn{2}{c}{CoCa} & \multicolumn{2}{c}{EVA-02} &\multicolumn{2}{c}{CLIP} & \multicolumn{2}{c}{SigLIP} \\
        \midrule
           Measure  & p-value     & $\tau$ & p-value     & $\tau$& p-value     & $\tau$& p-value     & $\tau$ \\
        \midrule
        \textbf{IIMM}          &  $<10^{-3}$ &   \textbf{0.89} &        0.001 &       \textbf{0.83} &       0.002 &      \textbf{0.78} &         0.002 &        0.78 \\
        \textbf{GBC} &  0.002 &   0.78 &        0.006 &       0.72 &       0.002 &      \textbf{0.78} &         0.001 &        \textbf{0.83} \\
        TransRate    &  0.013 &   0.67 &        0.025 &       0.61 &       0.013 &      0.67 &         0.006 &        0.72 \\
        EMMS          &  0.025 &   0.61 &        0.358 &       0.28 &       0.025 &      0.61 &         0.119 &        0.44 \\
        NCTI          &  0.919 &  -0.06 &        0.028 &      -0.59 &       0.818 &      0.07 &         0.827 &       -0.06 \\
        SFDA          &  0.260 &   0.33 &        0.919 &      -0.06 &       0.180 &      0.39 &         0.358 &        0.28 \\
        H-Score        &  0.761 &   0.11 &        0.002 &      -0.78 &       0.075 &      0.50 &         0.013 &       -0.67 \\
        \bottomrule
    \end{tabular}
    \caption{Kendall's $\tau$ correlation of score and gain over zero-shot error. Larger $\tau$ values are better.\sharedfootnotemark}
    \label{table:kendall gain over error}
\end{table*}

\begin{table*}[ht]
    \small
    \centering
    \begin{tabular}{lll|ll|ll|ll}
        \toprule
        \multicolumn{1}{c}{Model}& \multicolumn{2}{c}{CoCa} & \multicolumn{2}{c}{EVA-02} &\multicolumn{2}{c}{CLIP} & \multicolumn{2}{c}{SigLIP} \\
        \midrule
           Measure  & p-value     & $\tau$ & p-value     & $\tau$& p-value     & $\tau$& p-value     & $\tau$ \\
        \midrule
        \textbf{IIMM}          &  0.006 &  \textbf{0.72} &        0.006 &       \textbf{0.72} &       0.013 &      \textbf{0.67} &         0.025 &        0.61 \\
        \textbf{GBC} &  0.025 &  0.61 &        0.006 &       \textbf{0.72} &       0.013 &      \textbf{0.67} &         0.013 &       \textbf{0.67} \\
        TransRate     &  0.075 &  0.50 &        0.025 &       0.61 &       0.045 &      0.56 &         0.045 &        0.56 \\
        EMMS         &  0.013 &  0.67 &        0.025 &       0.61 &       0.075 &      0.50 &         0.025 &        0.61 \\
        NCTI          &  1.000 &  0.00 &        0.173 &      -0.37 &       0.818 &     -0.07 &         0.827 &        0.06 \\
        SFDA          &  0.075 &  0.50 &        0.612 &       0.17 &       0.075 &      0.50 &         0.260 &        0.33 \\
        H-Score       &  0.612 &  0.17 &        0.045 &      -0.56 &       0.025 &      0.61 &         0.075 &       -0.50\\
        \bottomrule
    \end{tabular}
    \caption{Kendall's $\tau$ correlation of score and accuracy. Larger $\tau$ values are better.\sharedfootnotemark}
    \label{table:kendall accuracy}
\end{table*}

\paragraph{Training details}
 To establish the ground truth for changes in model in-domain (ID) and out-of-domain (OOD) accuracy following fine-tuning on each dataset, we started by separately fine-tuning each of the base models on the nine datasets in our dataset suite. Following training on a given dataset, each model was tested on all nine datasets, with in-domain accuracy calculated as its accuracy on the test split of the training dataset and out-of-domain accuracies as the accuracies obtained on the remaining datasets test split. Each fine-tuning run consisted of 30 epochs of gradient descent using an SGD optimizer and a batch size of 128 and incorporated an exhaustive grid search over all relevant model hyper-parameters, examining values \{1e-2, 1e-3, 1e-4, 1e-5\} for the learning rate, \{1e-3, 1e-4, 1e-5\} for the weight decay, \{2, 4, 8, 16\} for LoRA model rank, and \{4, 8, 16, 32\} for CLIP-Adapter reduction. The initial image and text embeddings were then generated per dataset by a forward pass of each dataset's train split through the pre-trained base models and normalizing to unit vectors.

\subsection{Results}
\label{sec:results}

\begin{table*}[ht!]
    \small
    \centering
    \begin{tabular}{lll|ll|ll|ll}
        \toprule
        \multicolumn{1}{c}{Model}& \multicolumn{2}{c}{CoCa} & \multicolumn{2}{c}{EVA-02} &\multicolumn{2}{c}{CLIP} & \multicolumn{2}{c}{SigLIP} \\
        \midrule
        Measure  & p-value     & $R^2$ & p-value     & $R^2$& p-value     & $R^2$& p-value     & $R^2$ \\
        \midrule
        \textbf{IIMM}         &  $<10^{-3}$  &  \textbf{0.92} &        $<10^{-3}$  &      \textbf{0.89} &       $<10^{-3}$  &     \textbf{0.92} &         $<10^{-3}$  &       \textbf{0.92} \\
        GBC &  0.004 &  0.72 &        0.006 &      0.68 &       0.015 &     0.59 &         0.005 &       0.71 \\
        \bottomrule
    \end{tabular}
    \caption{Evaluation of linear fit of the IIMM to gain over zero-shot error and the GBC to the log of gain over zero-shot error. The significance of the slope coefficient is given by the p-value, and the strength of the fit is evaluated with the $R^2$ value.\sharedfootnotemark}
    \label{table:linreg gain over error}
\end{table*}

We first look at Kendall's $\tau$ correlation to determine the strength of the monotonic relationship of our proposed and comparison measures to both the fine-tuned gain over zero-shot error and fine-tuned accuracy.

The IIMM and the GBC have significantly stronger monotonic relationships than the other measures for both gain over zero-shot error, as shown in \cref{table:kendall gain over error}, and accuracy, as shown in \cref{table:kendall accuracy}. In both tables, the IIMM and the GBC are the most strongly correlated measures and display the greatest significance. Upon further investigation, the IIMM displays a mostly linear relationship with gain over zero-shot error, while the GBC displays a mostly exponential relationship with gain over zero-shot error. This allows us to fit and evaluate a predictive model using least-squares regression of the IIMM with gain over zero-shot error, and the GBC with the log of gain over zero-shot error. The results of these evaluations can be seen in \cref{table:linreg gain over error}, which shows the significance of the slope coefficient and the $R^2$ value, an estimate of the amount of variation within the data explained by its linear relationship to the measure. The IIMM has a stronger fit across all base pre-trained models compared to the GBC, and indeed shows itself to have a strong linear relationship to gain over zero-shot error with $R^2 > 0.89$ for all base models. Without the ability to capture cross-modal alignment, the compared uni-modal metrics expectedly demonstrate worse performance when compared to the IIMM, underscoring the value of an integrated vision-language metric.

\sharedfootnotetext

\begin{figure*}[ht]
  \centering
 \includegraphics[width=0.927\textwidth]{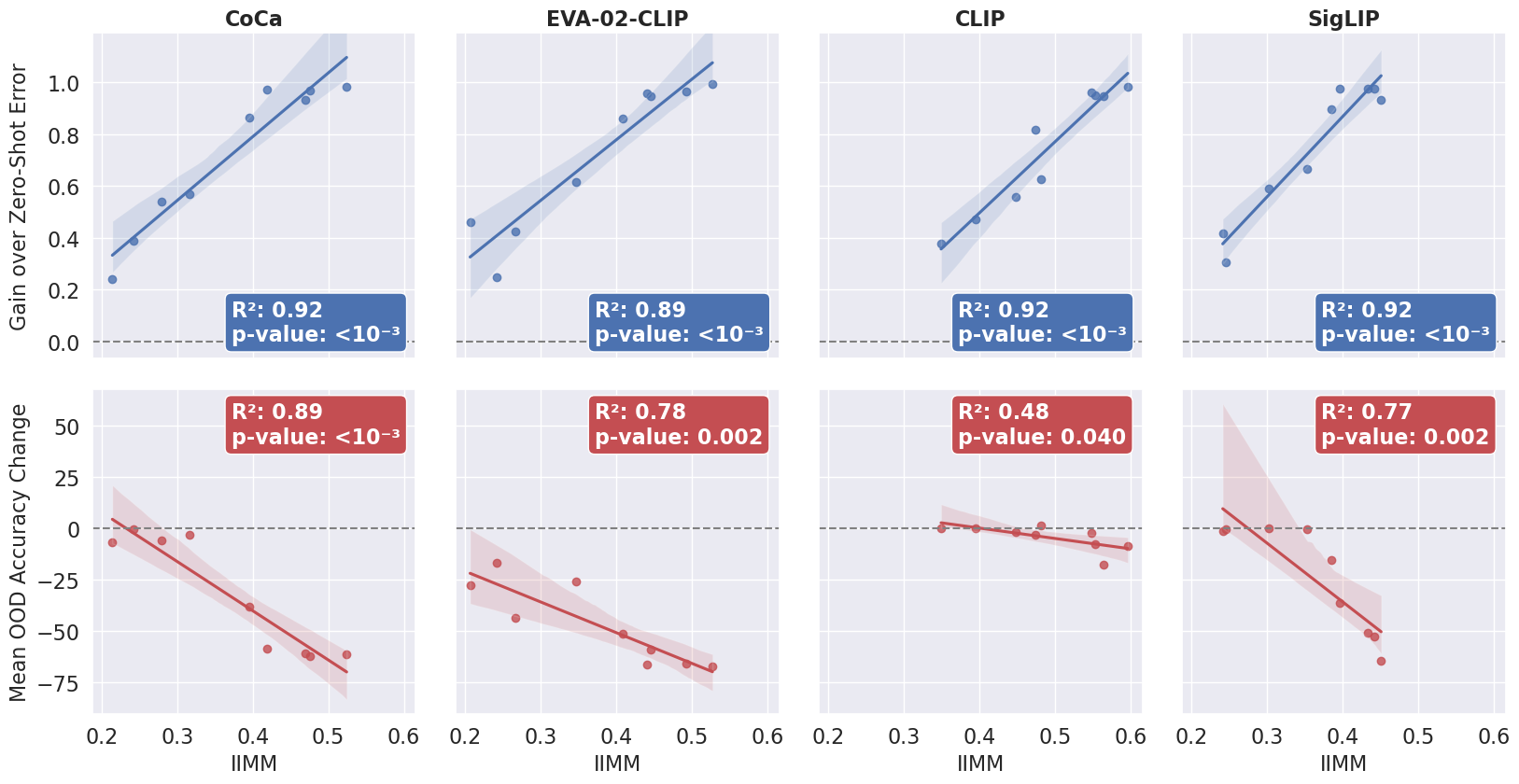}
  \caption{Performance gains and losses versus the IIMM per pre-trained model, with linear regression fit and 96\% confidence intervals. The top plots show the relationship between the on-target (ID) task IIMM to the on-target task performance gains over zero-shot error, while the bottom plots show the relationship between on-target task IIMM and the average performance loss on off-target (OOD) tasks.}
  \label{fig:IIMM by model}
\end{figure*}

\begin{figure*}[ht]
  \centering
    \begin{subfigure}[b]{1\textwidth}
         \centering
        \includegraphics[width=0.927\textwidth]{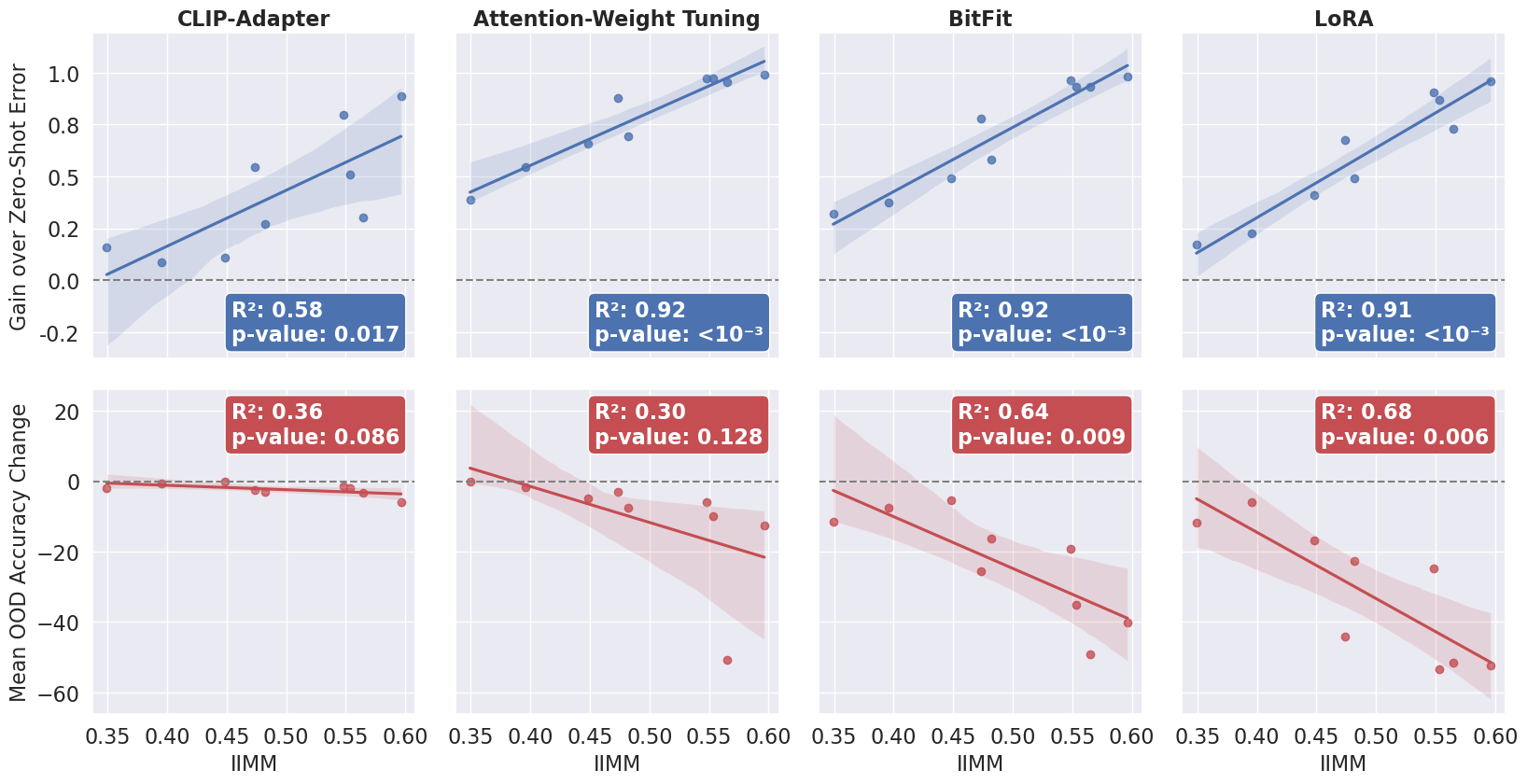}
 \end{subfigure}
  \caption{Performance gains and losses versus the IIMM per PEFT method, with linear regression fit and 96\% confidence intervals. The top plots show the relationship between the on-target task (ID) IIMM and on-target task performance gains over zero-shot error, while the bottom plots show the relationship between on-target task IIMM and the average performance loss on off-target (OOD) tasks.}
  \label{fig:peft}
\end{figure*}

This positive linear relationship between the IIMM and in-domain gain over zero-shot error is clearly seen in the top plots of \cref{fig:IIMM by model}. The bottom plots of \cref{fig:IIMM by model} show the negative linear relationship of the IIMM with the average change in accuracy across a suite of eight out-of-domain tasks. For all base pre-trained models aside from CLIP, the IIMM also shows a strong linear relationship to forgetting, as evidenced by the significance of the slope coefficient and high $R^2$ values. These relationships hold true across parameter-efficient fine-tuning methods, as shown in \cref{fig:peft}. We see with CLIP-Adapter that its overall worse in-domain performance attenuates the strength of the relationship between IIMM and gain over zero-shot error, and predictably, that a lower in-domain performance gain resulted in significantly less catastrophic forgetting. CLIP-Adapter’s design — injecting task kernels atop a frozen backbone — results in smaller, localized representational shifts, resulting in weaker in-distribution performance gains while preventing greater out-of-distribution performance loss. Since the IIMM is correlated with near-optimal performance gains, fine-tuning methods that result in sub-optimal gains may yield lower IIMM correlations.

These results provide evidence that the IIMM is strongly correlated with fine-tuning gains over zero-shot error. Unlike GBC, the second-highest correlated measure, the IIMM is bounded in [-1,1] and extreme values are informative in isolation.

\subsection{Exploring embedding features}
\label{sec:measure derivation}

\begin{table*}[!ht]
    \small
    \centering
    \begin{tabular}{lll|ll|ll|ll}
        \toprule
        \multicolumn{1}{c}{Model}& \multicolumn{2}{c}{CoCa} & \multicolumn{2}{c}{EVA-02} &\multicolumn{2}{c}{CLIP} & \multicolumn{2}{c}{SigLIP} \\
        \midrule
        Measure  & p-value     & $R^2$ & p-value     & $R^2$& p-value     & $R^2$& p-value     & $R^2$ \\
        \midrule
        \textbf{Incorrect Alignment} &  0.001& 0.82& 0.004& 0.72& 0.012& 0.62& 0.001& 0.80\\
        \textbf{Intra-Images Distance} & $<10^{-3}$ &0.88& $<10^{-3}$ &0.86& 0.001 &0.81& $<10^{-3}$ &0.88\\
        Intra-Texts Distance& 0.084& 0.37& 0.034& 0.50& 0.019& 0.57& 0.265& 0.17\\
        \bottomrule
    \end{tabular}
    \caption{Linear fit of measure to gain over zero-shot error with full fine-tuning by pre-trained model. The significance of the slope coefficient is given by the p-value, and the $R^2$ values represent the proportion of variation within the data modeled by the linear relationship.\sharedfootnotemark}
    \label{table:ar by model best measures}
\end{table*}

To empirically validate the provided theoretical justification for leveraging multi-modal information to improve transferability estimation in vision-language models, we independently evaluated all relevant components of CLIP's embedding geometry, known to manifest as a hypersphere. Using \eqref{eq:intra_modal}, we separately computed the average pairwise similarities for the image and text modalities and fit each to the accuracy gain over zero-shot error. We then applied \eqref{eq:inter_modal} to compute the degree of incorrect label alignment across the hypersphere and also fit the measure to the accuracy gain over zero-shot error. The final results of these evaluations can be seen in \cref{table:ar by model best measures}.

It is observed that the features of the embedding structure most predictive of accuracy gains are the \emph{intra-images distance} and the \emph{incorrect label alignment}. Based on these results, the correct label alignment and intra-texts distance are generally uninformative of accuracy gains with fine-tuning. As described in \cref{sec:IIMM}, each of these terms corresponds to a feature of the shared embedding space related to learning and forgetting. We include additional results of the Pearson correlations between different convex combinations of our inter- and intra- terms and gain over zero-shot error in the supplementary material. This empirical evidence successfully verifies that the terms combined have a stronger relationship with learning and forgetting than they do independently.

\section{Conclusion} \label{sec:conclusions}
In this work, we propose the Inter-Intra Modal Measure (IIMM) as a novel metric for predicting fine-tuning performance in vision-language models. We demonstrate that the IIMM strongly correlates with both learning gains and catastrophic forgetting, offering a simple yet powerful metric for estimating fine-tuning outcomes. Because the IIMM is bounded, an extreme value in isolation can inform practitioners whether a model will have limited or significant gains from fine-tuning. In addition, our results show that the IIMM can be leveraged to fit predictive linear models for estimating fine-tuning accuracy gains across diverse tasks with minimal computational overhead. We demonstrate that our proposed measure has a stronger relationship  to both fine-tuned accuracy and accuracy changes with fine-tuning than baselines. We further show the IIMM is informative of performance changes across parameter efficient fine-tuning methods, highlighting some methods are particularly susceptible to catastrophic forgetting when the zero-shot IIMM is high. Beyond empirical validation, we provide a theoretical bound linking changes in IIMM to the Wasserstein distance between pre- and post-fine-tuning embedding distributions, ensuring stability in IIMM predictions with respect to changes in representation space. Given the IIMM measures changes in embedding geometry prior to any task-specific heads, the metric is also task-agnostic, enabling its application to a variety of downstream tasks. We encourage future research to validate such use in broader applications. Overall, our work provides a practical and theoretically grounded tool for optimizing fine-tuning decisions in vision-language models, paving the way for more efficient and informed adaptation of foundation models to new tasks.

\paragraph{Limitations}
It is possible that the IIMM could show more variation when compared across a more diverse set of base models. While extreme values of the IIMM are heuristically informative in isolation, fitting a model for transfer accuracy estimation requires first fine-tuning across a small but diverse set of benchmarks. This could be an avenue of future research, as there might be additional information within the embedding geometry that allows for fine-tuning performance prediction without requiring information from fine-tuning on other tasks.

\section*{Acknowledgements}
The authors acknowledge the MIT Lincoln Laboratory Supercomputing Center for providing the high-performance computing resources that have contributed to the research results reported in this paper.

DISTRIBUTION STATEMENT A. Approved for public release. Distribution is unlimited. This material is based upon work supported by the Under Secretary of Defense for Research and Engineering under Air Force Contract No. FA8702-15-D-0001 or FA8702-25-D-B002. Any opinions, findings, conclusions or recommendations expressed in this material are those of the author(s) and do not necessarily reflect the views of the Under Secretary of Defense for Research and Engineering.





{
    \small
    \bibliographystyle{ieeenat_fullname}
    \bibliography{main}
}

\clearpage
\setcounter{page}{1}
\maketitlesupplementary

\section{Theorem \ref{thm:IIMM_bound} Proof} \label{appen:IIMM derivation}
\begin{proof}
  \textbf{Step 1. Representing IIMM as an Expectation}\\[1mm]
  The standard i.i.d. asumption on image embeddings yields $P_{XX}=P_X \otimes P_X$ and $Q_{XX}=Q_X \otimes Q_X$. By definition, we have
  \[
  \operatorname{IIMM}(P) = \frac{1}{2} \Bigl( \underbrace{\mathbb{E}_{(x,y') \sim P_{XY}}[x^\top y']}_{A(P)} + \underbrace{\mathbb{E}_{(x,x') \sim P_{XX}}[x^\top x']}_{B(P)} \Bigr).
  \]
  Similarly, with the perturbation,
  \[
  \operatorname{IIMM}(Q) = \frac{1}{2} \Bigl( \mathbb{E}_{(x,y') \sim Q_{XY}}[x^\top y'] + \mathbb{E}_{(x,x') \sim Q_{XX}}[x^\top x'] \Bigr).
  \]
  
  \textbf{Step 2. Lipschitz Continuity of the Inner Product}\\[1mm]
  Define the function 
  \[
  f(x,y) = x^\top y.
  \]
  For any two pairs $(x,y)$ and $(x',y')$ in $S^{d-1}\times S^{d-1}$, we have
  \[
  \begin{aligned}
  \bigl|f(x,y) - f(x',y')\bigr| &= \bigl| x^\top y - x'^\top y' \bigr| \\
  &\le \bigl|x^\top y - x'^\top y\bigr| + \bigl| x'^\top y - x'^\top y'\bigr| \\
  &\le \|x-x'\|_2\,\|y\|_2 + \|x'\|_2\,\|y-y'\|_2 \\
  &\le \|x-x'\|_2 + \|y-y'\|_2,
  \end{aligned}
  \]
  because $\|x\|_2 = \|y\|_2 = \|x'\|_2 = 1$. Hence, $f$ is $1$--Lipschitz with respect to the metric 
  \[
  d\bigl((x,y),(x',y')\bigr) = \|x-x'\|_2 + \|y-y'\|_2.
  \]
  
  \textbf{Step 3. Application of Kantorovich--Rubinstein Duality}\\[1mm]
  By the Kantorovich--Rubinstein duality, for any 1--Lipschitz function $f$ and any two probability measures $\mu$ and $\nu$, we have
  \[
  \Bigl|\mathbb{E}_{z\sim\mu}[f(z)] - \mathbb{E}_{z\sim\nu}[f(z)]\Bigr| \le W_1(\mu,\nu).
  \]
  Apply this result to the inter--modal term using $f(x,y)=x^\top y$. Thus,
  \begin{align*}
  &\Bigl| \mathbb{E}_{(x,y') \sim Q_{XY}}[x^\top y'] - \mathbb{E}_{(x,y') \sim P_{XY}}[x^\top y'] \Bigr| \\
  &\le W_1(P_{XY},Q_{XY}) \le \delta_A.
\end{align*}
  
  Similarly, for the intra--modal term, define 
  \[
  g(x,x') = x^\top x'.
  \]
  Using the same Lipschitz argument (with the metric $d((x,x'),(y,y')) = \|x-y\|_2 + \|x'-y'\|_2$) we obtain
\begin{align*}
  &\Bigl| \mathbb{E}_{(x,x') \sim Q_{XX}}[x^\top x'] - \mathbb{E}_{(x,x') \sim P_{XX}}[x^\top x'] \Bigr| \\
  &\le W_1(P_{XX},Q_{XX}) \le 2 W_1(P_X,Q_X) \le 2 \delta_B.
\end{align*}

  \textbf{Step 4. Combining the Two Contributions}\\[1mm]
  By the triangle inequality,
  \[
  \begin{aligned}
  &\Bigl| \operatorname{IIMM}(Q) - \operatorname{IIMM}(P) \Bigr|  \\
  &= \frac{1}{2}\Bigl| \bigl[A(Q)+B(Q)\bigr] - \bigl[A(P)+B(P)\bigr] \Bigr| \\
  &\le \frac{1}{2}\Bigl( \bigl| A(Q)-A(P) \bigr| + \bigl| B(Q)-B(P) \bigr| \Bigr) \\
  &\le \frac{1}{2} \bigl( \delta_A + 2 \delta_B \bigr).
  \end{aligned}
  \]
  
  \textbf{Step 5. Finite--Sample Concentration}\\[1mm]
  In practice, one estimates the expectations in $A(P)$ and $B(P)$ from $N$ independent samples. Since the inner product $x^\top y$ is bounded in $[-1,1]$, standard concentration inequalities (e.g., Hoeffding's inequality) imply that, with probability at least $1-\eta$, the empirical estimates $\widehat{A}(P)$ and $\widehat{B}(P)$ satisfy
  \[
  \Bigl| \widehat{A}(P) - A(P) \Bigr| \le \epsilon_N \quad \text{and} \quad \Bigl| \widehat{B}(P) - B(P) \Bigr| \le \epsilon_N,
  \]
  where
  \[
  \epsilon_N = O\!\Bigl(\sqrt{\frac{\log(1/\eta)}{N}}\Bigr).
  \]
  Hence, with the corresponding estimates $\widehat{\operatorname{IIMM}}(P)$ and $\widehat{\operatorname{IIMM}}(Q)$, we have
  \[
  \Bigl| \widehat{\operatorname{IIMM}}(Q) - \widehat{\operatorname{IIMM}}(P) \Bigr| \le \frac{\delta_A + 2 \delta_B}{2} + 2\epsilon_N,
  \]
  where the constant $2$ in front of $\epsilon_N$ can be absorbed in the big-$O$ notation.
  
  This completes the proof.
\end{proof}

\section{Combining inter- and intra- measures}
\label{append:cc inter and intra}
To determine how best to combine the inter- and intra- measures, we used the Pearson correlation, $r_p$, and its $95\%$ CI of the measure determined by different convex combinations of the inter- and intra- measure and the gain over zero-shot error. We see in \Cref{tab:cc base} and \Cref{tab:cc peft} that we do not have enough power to make strong conclusions about the best coefficient value. We do see a pattern of smaller variance in the correlation estimate when combining the inter- and intra- terms compared to using either term in isolation. 

We chose a coefficient of $0.5$ for simplicity in further analysis.

\section{Additional Results}
To ensure completeness and reproducibility, \Cref{tab:base-model-performance} and \Cref{tab:peft-model-performance} present the per-task zero-shot accuracy, fine-tuned accuracy, raw accuracy gain, and gain over zero-shot error for each base model and PEFT method, respectively. Kendall's tau scores, detailed in \Cref{table:kendall gain over error} and \Cref{table:kendall accuracy} of the main text, were calculated by first computing the transferability measures of interest for each dataset's zero-shot embeddings. These transferability measures, combined with the per-dataset gains over zero-shot error and raw accuracy gains, form paired observations across datasets. Kendall’s tau was then computed using:
\[
\tau = \frac{P - Q}{\sqrt{(P + Q + T)(P + Q + U)}}
\]
Here, $P$ and $Q$ represent the number of concordant and discordant pairs, respectively. The formula considers all possible pairs of datasets, comparing their relative ordering in each list. Ties are corrected for by $T$ and $U$, which denote the number of tied pairs in the first and second list, respectively.

\begin{table*}[hb]
    \centering
    \caption{Performance metrics per model and dataset. ZS Acc: Zero-Shot Accuracy, FT Acc: Fine-Tuned Accuracy, Raw Gain: FT Acc - ZS Acc, Gain over ZS: Normalized Gain over Zero-Shot Error.}
    \label{tab:base-model-performance}
    \begin{tabular}{llcccc}
        \toprule
        Model & Dataset & ZS Acc (\%) & FT Acc (\%) & Raw Gain (\%) & Gain over ZS \\
        \midrule
        CLIP & Cars & 58.87 & 78.26 & 19.39 & 0.47 \\
         & DTD & 42.07 & 74.31 & 32.23 & 0.56 \\
         & SVHN & 27.27 & 96.00 & 68.73 & 0.95 \\
         & EuroSAT & 44.07 & 97.85 & 53.78 & 0.96 \\
         & CIFAR100 & 61.71 & 85.67 & 23.96 & 0.63 \\
         & GTSRB & 33.65 & 96.72 & 63.07 & 0.95 \\
         & MNIST & 50.47 & 99.23 & 48.76 & 0.98 \\
         & RESISC45 & 56.56 & 92.10 & 35.54 & 0.82 \\
         & SUN397 & 61.34 & 75.95 & 14.61 & 0.38 \\
         & STL10 & 97.36 & 98.42 & 1.06 & 0.40 \\
        \midrule
        CoCa & Cars & 83.85 & 90.10 & 6.26 & 0.39 \\
         & DTD & 51.91 & 79.26 & 27.34 & 0.57 \\
         & SVHN & 54.74 & 96.90 & 42.16 & 0.93 \\
         & EuroSAT & 42.85 & 98.26 & 55.41 & 0.97 \\
         & CIFAR100 & 74.12 & 88.07 & 13.95 & 0.54 \\
         & GTSRB & 42.39 & 98.30 & 55.91 & 0.97 \\
         & MNIST & 69.04 & 99.45 & 30.41 & 0.98 \\
         & RESISC45 & 60.13 & 94.60 & 34.48 & 0.86 \\
         & SUN397 & 66.24 & 74.40 & 8.17 & 0.24 \\
         & STL10 & 96.24 & 98.06 & 1.83 & 0.49 \\
        \midrule
        EVA-02 & Cars & 78.80 & 88.55 & 9.75 & 0.46 \\
         & DTD & 50.96 & 81.22 & 30.27 & 0.62 \\
         & SVHN & 24.92 & 97.26 & 72.34 & 0.96 \\
         & EuroSAT & 68.04 & 98.33 & 30.30 & 0.95 \\
         & CIFAR100 & 87.64 & 92.90 & 5.26 & 0.43 \\
         & GTSRB & 46.64 & 97.78 & 51.14 & 0.96 \\
         & MNIST & 44.21 & 99.63 & 55.42 & 0.99 \\
         & RESISC45 & 65.57 & 95.22 & 29.65 & 0.86 \\
         & SUN397 & 70.71 & 78.02 & 7.31 & 0.25 \\
         & STL10 & 99.48 & 99.51 & 0.04 & 0.07 \\
        \midrule
        SigLIP & Cars & 90.80 & 94.64 & 3.84 & 0.42 \\
         & DTD & 62.61 & 84.63 & 22.02 & 0.59 \\
         & SVHN & 55.87 & 96.95 & 41.07 & 0.93 \\
         & EuroSAT & 43.63 & 98.63 & 55.00 & 0.98 \\
         & CIFAR100 & 70.91 & 90.28 & 19.37 & 0.67 \\
         & GTSRB & 52.84 & 98.85 & 46.01 & 0.98 \\
         & MNIST & 83.52 & 99.62 & 16.10 & 0.98 \\
         & RESISC45 & 60.56 & 95.89 & 35.33 & 0.90 \\
         & SUN397 & 70.23 & 79.31 & 9.08 & 0.31 \\
         & STL10 & 98.19 & 99.18 & 0.99 & 0.54 \\
        \bottomrule
    \end{tabular}
\end{table*}

\begin{table*}[hb]
    \centering
    \caption{Performance metrics per PEFT method and dataset. ZS Acc: Zero-Shot Accuracy, FT Acc: Fine-Tuned Accuracy, Raw Gain: FT Acc - ZS Acc, Gain over ZS: Normalized Gain over Zero-Shot Error.}
    \label{tab:peft-model-performance}
    \begin{tabular}{llcccc}
        \toprule
        Model & Dataset & ZS Acc (\%) & FT Acc (\%) & Raw Gain (\%) & Gain over ZS \\
        \midrule
        Attention-Weight Tuning & Cars & 58.87 & 81.26 & 22.39 & 0.54 \\
         & DTD & 42.07 & 80.16 & 38.09 & 0.66 \\
         & SVHN & 27.27 & 96.57 & 69.29 & 0.95 \\
         & EuroSAT & 44.07 & 98.52 & 54.44 & 0.97 \\
         & CIFAR100 & 61.71 & 88.16 & 26.45 & 0.69 \\
         & GTSRB & 33.65 & 98.15 & 64.50 & 0.97 \\
         & MNIST & 50.47 & 99.56 & 49.09 & 0.99 \\
         & RESISC45 & 56.56 & 94.62 & 38.06 & 0.88 \\
         & SUN397 & 61.34 & 76.31 & 14.97 & 0.39 \\
         & STL10 & 97.36 & 98.44 & 1.08 & 0.41 \\
        \midrule
        BitFit & Cars & 58.87 & 74.16 & 15.28 & 0.37 \\
         & DTD & 42.07 & 70.43 & 28.35 & 0.49 \\
         & SVHN & 27.27 & 95.15 & 67.88 & 0.93 \\
         & EuroSAT & 44.07 & 98.04 & 53.96 & 0.96 \\
         & CIFAR100 & 61.71 & 83.90 & 22.19 & 0.58 \\
         & GTSRB & 33.65 & 95.30 & 61.65 & 0.93 \\
         & MNIST & 50.47 & 99.15 & 48.68 & 0.98 \\
         & RESISC45 & 56.56 & 90.37 & 33.81 & 0.78 \\
         & SUN397 & 61.34 & 73.68 & 12.34 & 0.32 \\
         & STL10 & 97.36 & 98.31 & 0.95 & 0.36 \\
        \midrule
        LoRA & Cars & 58.87 & 68.23 & 9.35 & 0.23 \\
         & DTD & 42.07 & 65.90 & 23.83 & 0.41 \\
         & SVHN & 27.27 & 80.34 & 53.07 & 0.73 \\
         & EuroSAT & 44.07 & 94.63 & 50.56 & 0.90 \\
         & CIFAR100 & 61.71 & 80.49 & 18.78 & 0.49 \\
         & GTSRB & 33.65 & 91.39 & 57.74 & 0.87 \\
         & MNIST & 50.47 & 97.89 & 47.42 & 0.96 \\
         & RESISC45 & 56.56 & 85.87 & 29.32 & 0.67 \\
         & SUN397 & 61.34 & 67.99 & 6.65 & 0.17 \\
         & STL10 & 97.36 & 95.03 & -2.34 & -0.89 \\
        \midrule
        CLIP-Adapter & Cars & 58.87 & 62.45 & 3.58 & 0.09 \\
         & DTD & 42.07 & 48.24 & 6.17 & 0.11 \\
         & SVHN & 27.27 & 49.30 & 22.03 & 0.30 \\
         & EuroSAT & 44.07 & 88.52 & 44.44 & 0.79 \\
         & CIFAR100 & 61.71 & 72.13 & 10.42 & 0.27 \\
         & GTSRB & 33.65 & 67.32 & 33.67 & 0.51 \\
         & MNIST & 50.47 & 94.26 & 43.79 & 0.88 \\
         & RESISC45 & 56.56 & 80.21 & 23.65 & 0.54 \\
         & SUN397 & 61.34 & 67.46 & 6.12 & 0.16 \\
         & STL10 & 97.36 & 95.03 & -2.34 & -0.89 \\
        \bottomrule
    \end{tabular}
\end{table*}

\begin{table*}[hb]
    \centering
    \caption{Pearson correlation and 95\% confidence intervals of zero-shot measure and gain over zero-shot error for different values of $\alpha$ in convex combinations of the inter and intra measures, $\alpha\text{intra} + (1-\alpha)\text{inter}$. Data from four base models trained over 9 tasks.}
    \label{tab:cc base}
    \begin{tabular}{lllllllll}
         \toprule\\
         \multicolumn{1}{c}{Model}& \multicolumn{2}{c}{CLIP} & \multicolumn{2}{c}{CoCa} &\multicolumn{2}{c}{EVA-02} & \multicolumn{2}{c}{SigLIP} \\
         \midrule
         $\alpha$ & $r_p$ & 95\% CI & $r_p$ & 95\% CI & $r_p$ & 95\% CI & $r_p$ & 95\% CI \\
         \midrule
           0.00 &  0.79& (0.26, 0.95) &  0.91& (0.62, 0.98) &  0.85& (0.43, 0.97) &  0.89& (0.55, 0.98) \\
           0.10 &  0.92& (0.66, 0.98) &  0.95& (0.77, 0.99) &  0.91& (0.62, 0.98) &  0.94& (0.73, 0.99) \\
           0.20 &  0.97& (0.86, 0.99) &  0.96& (0.82, 0.99) &  0.94& (0.73, 0.99) &  0.96& (0.82, 0.99) \\
           0.30 &    0.98& (0.9, 1.0) &  0.97& (0.86, 0.99) &  0.95& (0.77, 0.99) &  0.97& (0.86, 0.99) \\
           0.40 &  0.97& (0.86, 0.99) &  0.97& (0.86, 0.99) &  0.95& (0.77, 0.99) &  0.96& (0.82, 0.99) \\
           0.50 &  0.96& (0.82, 0.99) &  0.96& (0.82, 0.99) &  0.95& (0.77, 0.99) &  0.96& (0.82, 0.99) \\
           0.60 &  0.95& (0.77, 0.99) &  0.96& (0.82, 0.99) &  0.94& (0.73, 0.99) &  0.96& (0.82, 0.99) \\
           0.70 &   0.93& (0.7, 0.99) &  0.95& (0.77, 0.99) &  0.94& (0.73, 0.99) &  0.95& (0.77, 0.99) \\
           0.80 &  0.92& (0.66, 0.98) &  0.95& (0.77, 0.99) &   0.93& (0.7, 0.99) &  0.95& (0.77, 0.99) \\
           0.90 &  0.91& (0.62, 0.98) &  0.94& (0.73, 0.99) &   0.93& (0.7, 0.99) &  0.94& (0.73, 0.99) \\
           1.00 &  0.90& (0.59, 0.98) &  0.94& (0.73, 0.99) &   0.93& (0.7, 0.99) &  0.94& (0.73, 0.99) \\
         \bottomrule
    \end{tabular}
\end{table*}

\begin{table*}[hb]
    \centering
    \caption{Pearson correlation and 95\% confidence intervals of zero-shot measure and gain over zero-shot error for different values of $\alpha$ in convex combinations of the inter and intra measures, $\alpha\text{intra} + (1-\alpha)\text{inter}$. Data from PEFT methods trained over 9 tasks.}
    \label{tab:cc peft}
    \begin{tabular}{lllllllll}
         \toprule\\
         \multicolumn{1}{c}{Method}& \multicolumn{2}{c}{Attention-WT} & \multicolumn{2}{c}{BitFit} &\multicolumn{2}{c}{LoRA} & \multicolumn{2}{c}{Adapter} \\
         \midrule
         $\alpha$ & $r_p$ & 95\% CI & $r_p$ & 95\% CI & $r_p$ & 95\% CI & $r_p$ & 95\% CI\\
         \midrule
          0.00 & 0.81 & (0.32, 0.96) & 0.78 & (0.24, 0.95) & 0.80 & (0.29, 0.96) & 0.61 & (-0.09, 0.91) \\
          0.10 &  0.93 & (0.7, 0.99) & 0.92 & (0.66, 0.98) & 0.92 & (0.66, 0.98) &  0.72 & (0.11, 0.94) \\
          0.20 &   0.98 & (0.9, 1.0) & 0.97 & (0.86, 0.99) & 0.97 & (0.86, 0.99) &  0.76 & (0.19, 0.95) \\
          0.30 &   0.98 & (0.9, 1.0) &   0.98 & (0.9, 1.0) &   0.98 & (0.9, 1.0) &  0.77 & (0.22, 0.95) \\
          0.40 & 0.97 & (0.86, 0.99) & 0.97 & (0.86, 0.99) & 0.97 & (0.86, 0.99) &  0.77 & (0.22, 0.95) \\
          0.50 & 0.96 & (0.82, 0.99) & 0.96 & (0.82, 0.99) & 0.95 & (0.77, 0.99) &  0.76 & (0.19, 0.95) \\
          0.60 & 0.94 & (0.73, 0.99) & 0.94 & (0.73, 0.99) & 0.94 & (0.73, 0.99) &  0.75 & (0.17, 0.94) \\
          0.70 &  0.93 & (0.7, 0.99) &  0.93 & (0.7, 0.99) &  0.93 & (0.7, 0.99) &  0.74 & (0.15, 0.94) \\
          0.80 & 0.92 & (0.66, 0.98) & 0.92 & (0.66, 0.98) & 0.91 & (0.62, 0.98) &  0.73 & (0.13, 0.94) \\
          0.90 & 0.90 & (0.59, 0.98) & 0.91 & (0.62, 0.98) & 0.90 & (0.59, 0.98) &  0.73 & (0.13, 0.94) \\
          1.00 & 0.89 & (0.55, 0.98) & 0.90 & (0.59, 0.98) & 0.89 & (0.55, 0.98) &  0.72 & (0.11, 0.94)\\
         \bottomrule
    \end{tabular}
\end{table*}

\begin{table*}[ht]
  \caption{Linear fit of IIMM with different intra-modal measures to gain over zero-shot error following fine-tuning by pre-trained model (alpha = 0.5).\sharedfootnotemark}
  \label{table:iimm_alternatives}
  \centering
  \begin{tabular}{lll|ll|ll|ll}
    \toprule
    \multicolumn{1}{c}{Model}& \multicolumn{2}{c}{CoCa} & \multicolumn{2}{c}{EVA-02} &\multicolumn{2}{c}{CLIP} & \multicolumn{2}{c}{SigLIP} \\
    \midrule
       Intra-Modal Measure  & p-value     & $r_s$ & p-value     & $r_s$& p-value     & $r_s$& p-value     & $r_s$ \\
    \midrule
    \textbf{Intra-Images Distance} &$\mathbf{<10^{-3}}$ &\textbf{0.95}& $\mathbf{<10^{-3}} $&\textbf{0.93}& \textbf{0.001} &\textbf{0.91}& \textbf{0.002} &\textbf{0.88}\\
    H-Score & 0.020& 0.75 & 0.077& 0.62 & 0.002& 0.88 & 0.020& 0.75 \\
    TransRate &	0.004& 0.85& 0.007& 0.82& 0.005& 0.83& 0.004& 0.85\\
    GBC & 0.002& 0.88& 0.050& 0.67& 0.042 &0.68& 0.016& 0.77\\
    \bottomrule
  \end{tabular}
\end{table*}

\end{document}